\documentclass[11pt]{article}

\usepackage[T1]{fontenc}
\usepackage[utf8]{inputenc}
\usepackage{lmodern}
\usepackage{microtype}
\usepackage{geometry}
\usepackage{enumitem}
\usepackage{amsmath}
\usepackage{amssymb}
\usepackage{hyperref}
\usepackage{graphicx}
\usepackage{svg}

\usepackage{fvextra}
\DefineVerbatimEnvironment{KapsoAlgo}{Verbatim}{
  fontsize=\small,
  breaklines=true,
  breakanywhere=true,
  breaksymbolleft={}
}

\title{KAPSO: A Knowledge-grounded framework for Autonomous Program Synthesis and Optimization}

\author{
    Alireza Nadafian\thanks{Equal contribution} \quad 
    Alireza Mohammadshahi\footnotemark[1] \quad 
    Majid Yazdani\footnotemark[1] \\
    \text{Leeroo Team} \\
    \texttt{\{nadaf, alireza, my\}@leeroo.com}
}

\date{}

\begin{document}
\maketitle

\begin{abstract}
We introduce \textbf{KAPSO}, a modular framework for \textbf{autonomous program synthesis and optimization}. Given a natural-language goal and an evaluation method, KAPSO iteratively performs \textbf{ideation}, \textbf{code synthesis and editing}, \textbf{execution}, \textbf{evaluation}, and \textbf{learning} to improve a runnable artifact toward measurable objectives. Rather than treating synthesis as the endpoint, KAPSO uses synthesis as an operator within a long-horizon optimization loop, where progress is defined by evaluator outcomes.

KAPSO targets long-horizon failures common in coding agents—including lost experimental state, brittle debugging, and weak reuse of domain expertise—by integrating three tightly coupled components. First, a \textbf{git-native experimentation engine} isolates each attempt as a branch, producing reproducible artifacts and preserving provenance across iterations. Second, a \textbf{knowledge system} ingests heterogeneous sources, including repositories, internal playbooks, and curated external resources such as documentation, scientific papers, and web search results, and organizes them into a structured representation that supports retrieval over workflows, implementations, and environment constraints. Third, a \textbf{cognitive memory layer} coordinates retrieval and maintains an \textbf{episodic store} of reusable lessons distilled from experiment traces (run logs, diffs, and evaluator feedback), reducing repeated error modes and accelerating convergence.

We evaluated KAPSO on \textbf{MLE-Bench} (Kaggle-style ML competitions) and \textbf{ALE-Bench} (AtCoder heuristic optimization), and report end-to-end performance.
\end{abstract}

\section{Introduction}

Domain experts often know what they want to build, but turning that intent into reliable, runnable, and optimized software still requires repeated experimentation. In practice, successful development is an iterative process: propose an approach, implement it, run it in the real environment, inspect outcomes, and refine. This loop is especially visible in data and AI programs, where progress depends on many measurable improvements and on careful management of code, data, and evaluation contracts. Importantly, iterations often succeed in the narrow sense of producing a working artifact, yet still fall short on quality, accuracy, robustness, or efficiency. Practical progress therefore requires repeated evaluation and targeted improvement, not only error fixing.

LLM-based coding agents reduce the cost of writing code, but they remain unreliable in long-horizon execution loops. Common failure modes include losing state across iterations, repeatedly triggering the same integration errors, and failing to reuse relevant engineering expertise even when it is available in repositories, documentation, internal playbooks, or prior attempts. In many real settings, the decisive advantage is not raw code generation, but the ability to consistently apply expert-grade ideas and high-leverage engineering workflows, including environment setup, data contracts, evaluation harnesses, debugging procedures, and performance tuning. 

We present \textbf{KAPSO}, a framework for \textbf{execution-grounded program optimization} under an explicit evaluator boundary. Given a natural-language goal and evaluators, KAPSO runs an iterative \textbf{evolve loop}: it generates and selects improvement hypotheses, synthesizes and applies code edits, executes the resulting artifact, evaluates outcomes, and uses measured feedback to guide subsequent iterations. In KAPSO, \emph{program synthesis is not the endpoint}; it is an operator within a long-horizon optimization process where progress is defined by evaluator outcomes such as accuracy, robustness, efficiency, or preference-based quality.

KAPSO integrates three tightly coupled components to make this optimization loop reliable and reusable. First, a \textbf{git-native experimentation engine} isolates each attempt as a branch, capturing code changes, logs, and evaluation outputs as reproducible artifacts with explicit provenance. Second, a \textbf{knowledge system} ingests heterogeneous sources, including repositories, benchmark artifacts, internal playbooks, documentation, scientific papers, and web-derived material, and organizes them into a structured representation that supports retrieval of ideas, implementations, heuristics, and environment constraints. This knowledge is hosted in \textbf{MediaWiki}, providing a familiar interface for human review, curation, and human-in-the-loop iteration. We release a complete knowledge package consisting of a MediaWiki dump, Neo4j and Weaviate snapshots, and Docker-based deployment scripts that bring up the MediaWiki instance and all indices in a reproducible configuration. Third, a \textbf{cognitive memory layer} coordinates knowledge retrieval from the knowledge base and maintains an episodic store of reusable lessons distilled from experiment traces (run logs, diffs, and evaluator feedback), reducing repeated error modes and accelerating convergence.

KAPSO is intentionally modular. It supports pluggable evaluators, knowledge backends, and coding agents, enabling the same optimization loop to be applied across domains where progress is defined by executable outcomes and measurable objectives. We instantiate this design and evaluate it on two complementary benchmarks, MLE-Bench and ALE-Bench, and we use these instantiations to study end-to-end performance. Beyond these benchmarks, the same interfaces generalize to additional tasks by swapping the evaluator and the knowledge sources.

\paragraph{Contributions.}
This paper makes the following contributions:
\begin{enumerate}[leftmargin=1.2em]
  \item An end-to-end framework for \textbf{evaluator-grounded program optimization} that improves runnable artifacts through iterative ideation, code synthesis/editing, execution, evaluation, and learning.
  \item A git-native experimentation engine that represents each attempt as an isolated, reproducible branch with explicit provenance.
  \item A knowledge acquisition and representation pipeline hosted in MediaWiki that converts heterogeneous sources into a typed, workflow-oriented knowledge base usable during optimization.
  \item A cognitive memory system that combines knowledge retrieval with episodic learning from experiment traces to reduce repeated failures and accelerate iteration.
  \item A modular architecture with pluggable evaluators and knowledge sources, demonstrated through benchmark instantiations and ablations, together with a released knowledge package containing a MediaWiki dump, Neo4j and Weaviate snapshots, and Docker-based deployment scripts for reproducing the full stack. The released knowledge base is populated from over 2{,}000 widely used data and ML repositories, with selection criteria defined later.
\end{enumerate}

\section{Framework Overview}

KAPSO is designed around a simple contract: given a natural-language goal, optional knowledge sources, and an evaluator, the system produces a runnable software artifact and improves it through measured iteration. In typical use, users interact with KAPSO through two core operations, \texttt{evolve} and \texttt{deploy}, and optionally use \texttt{learn} and \texttt{research} to expand or refresh the knowledge base.

KAPSO exposes a user-facing \texttt{Kapso} API with four operations:

\begin{itemize}[leftmargin=1.2em]
  \item \textbf{\texttt{evolve}}: run an evaluator-grounded optimization loop that proposes improvements, applies code edits (including synthesis when needed), executes the artifact, and uses measured outcomes to guide subsequent iterations.
  \item \textbf{\texttt{deploy}}: adapt a selected solution into a target runtime strategy and return a unified runtime handle.
  \item \textbf{\texttt{learn}}: ingest and curate user-provided knowledge sources (e.g., repositories, internal playbooks, benchmark artifacts) into a unified knowledge base with both human-facing and retrieval-facing representations.
  \item \textbf{\texttt{research}} : discover domain-relevant external material (e.g., papers, documentation, web-derived notes, and candidate repositories) and return structured findings that can be used as context for \texttt{evolve} and as seeds for \texttt{learn}.
\end{itemize}

In particular, \texttt{learn} is not required when a suitable knowledge base already exists, and \texttt{research} is not required when the user already knows which sources to ingest. When sources are unknown or incomplete, \texttt{research} can be used to identify candidate materials, after which \texttt{learn} incorporates selected sources into the knowledge system for reuse across tasks.

\begin{figure}[h]
\centering
\includegraphics[width=0.95\linewidth]{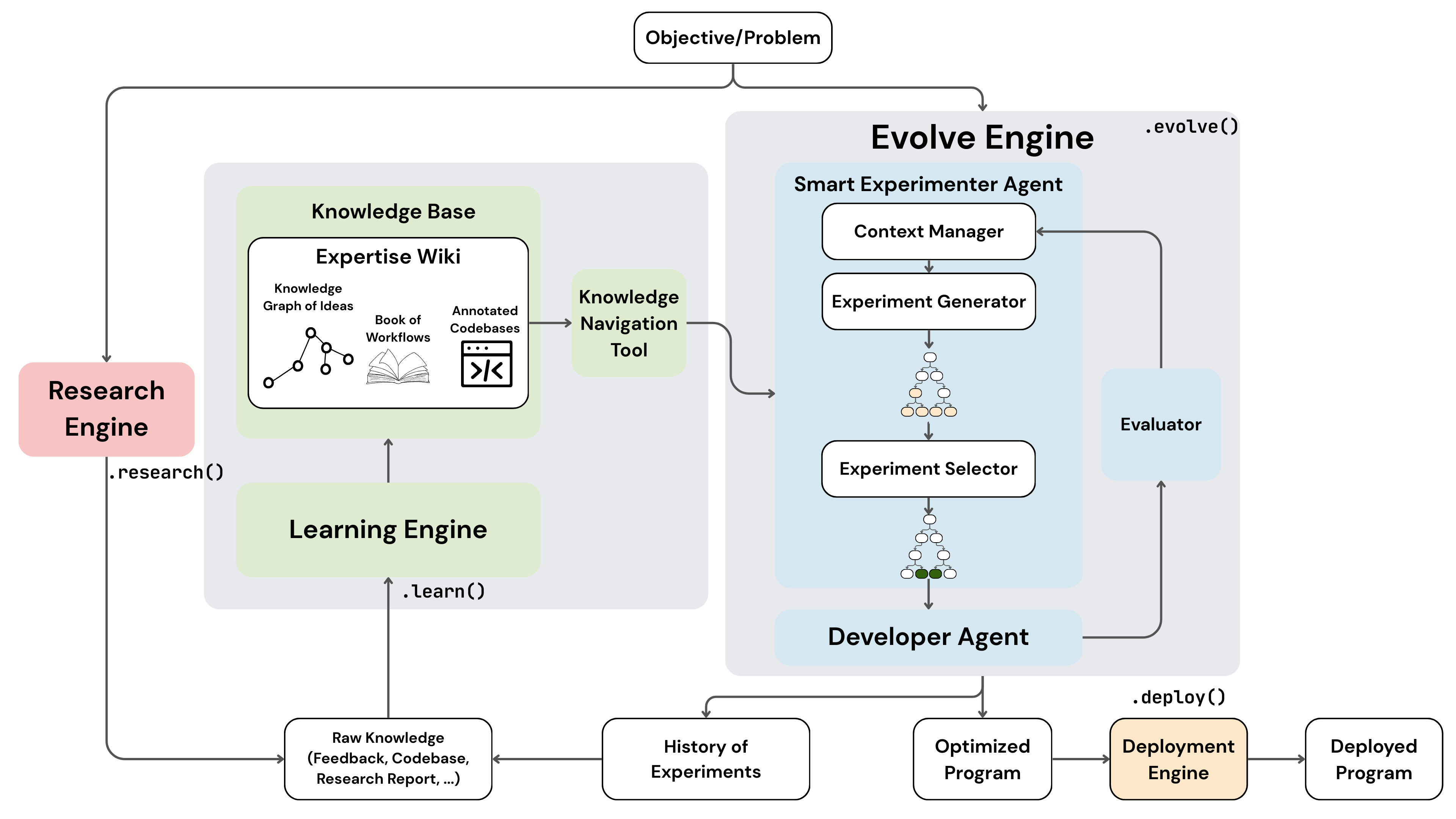}
\caption{The overview of the framework architecture.}
\label{fig:framework_overview}
\end{figure}

\subsection{Knowledge plane vs.\ execution plane}

KAPSO separates \emph{knowledge} from \emph{execution}. The knowledge plane aggregates heterogeneous sources such as repositories, documentation, scientific papers, web-derived material, benchmark artifacts, and internal playbooks. It is hosted in \textbf{MediaWiki} for human review and curation, and it is indexed into \textbf{pluggable retrieval backends} (for example, graph and vector indices) to support machine retrieval at evolve time. In addition, KAPSO generates self-knowledge from its own work by extracting lessons from experiment traces and storing them in an \textbf{episodic memory} store. Implementation details of ingestion, indexing, and the released knowledge package are described in Section~\ref{sec:implementation:knowledge}.

The execution plane is anchored by an explicit evaluator boundary. A \texttt{ProblemHandler} (and optionally an \texttt{Evaluator}) defines the task contract: how the artifact is executed, what outputs must be produced, and how quality is measured. Evaluators may be fully automated (metrics, tests, graders), may be stochastic, and may incorporate LLM-based judges or qualitative preference rules rather than a single scalar objective. This boundary allows the same evolve loop to generalize across domains by swapping the evaluator while keeping orchestration logic unchanged.

\subsection{Evolve orchestration and modular subsystems}

Internally, \texttt{Kapso.evolve} is implemented by an \texttt{OrchestratorAgent} that composes four pluggable subsystems. These subsystems are configured per task and can be replaced independently:

\begin{itemize}[leftmargin=1.2em]
  \item \textbf{SearchStrategy}: proposes candidate solution specifications and selects which candidates to evaluate next. Concrete strategies (for example, linear search and tree-based search) are defined formally later in the paper.
  \item \textbf{ContextManager}: renders the context passed into solution proposal and implementation, including the task description, constraints, retrieved knowledge, episodic insights, and summaries of prior experiments.
  \item \textbf{KnowledgeSearch}: retrieves workflows, implementations, heuristics, and environment constraints from the knowledge base, and returns structured knowledge packets with provenance.
  \item \textbf{CodingAgent}: applies code changes to implement or debug a candidate solution.
\end{itemize}

The \texttt{OrchestratorAgent} drives an iterative loop that alternates between (i) constructing context from evaluator state, knowledge retrieval, episodic memory, and experiment history, and (ii) executing one or more isolated experiments to obtain measured outcomes.

\subsection{Deploy and the unified runtime interface}
\label{sec:framework:deploy}

Today, \texttt{deploy} returns a Python \texttt{Software} handle with a unified \texttt{run()} interface, backed by an adapted copy of the solution repository. The \texttt{Software} handle exposes \texttt{run(inputs)} and returns a normalized dictionary (for example, \texttt{\{"status": "success", "output": ...\}} or \texttt{\{"status": "error", "error": ...\}}), plus lifecycle methods such as \texttt{start}, \texttt{stop}, \texttt{logs}, and \texttt{is\_healthy}. Depending on the selected strategy it executes locally (import-and-call) or via an HTTP or remote endpoint (for example, Docker, Modal, BentoCloud, or LangGraph Platform), while preserving the same \texttt{run()} contract for callers. Implementation details of repository adaptation and strategy adapters are described in Section~\ref{sec:implementation:deploy}.

\subsection{Experiment artifacts and provenance}

KAPSO represents each experiment as an isolated branch and persists run artifacts sufficient to reproduce, debug, and reuse successful attempts. The experiment artifact model and branch publication behavior are implemented by the experimentation engine described in Section~\ref{sec:implementation:experiments}.

\section{Formalization (Notation and Algorithms)}
\label{sec:formalization}

This section introduces notation and formal operators for the KAPSO framework. The goal is to make the system comparable to prior work in program synthesis, agentic search, and black-box optimization by stating (i) the objects KAPSO manipulates, (ii) the evaluation contract it relies on, and (iii) the algorithms it runs. Benchmarks such as MLE-Bench and ALE-Bench are treated later as concrete instantiations of the evaluator contract.

\subsection{Evolve instance and evaluator contract}

A KAPSO run is defined by a natural-language goal $g$, a budget specification $B$ (time, iterations, and/or cost), and an evaluator contract $\mathcal{E}$. The evaluator contract defines how artifacts are executed, measured, compared, and when the run should stop:

\begin{itemize}[leftmargin=1.2em]
  \item \textbf{Budget progress}: $\beta_i \in [0,1]$ is the normalized budget progress at outer iteration $i$.
  \item \textbf{Problem context}: $P(\beta)$ returns the context shown to the system at budget progress $\beta$ (optionally budget-aware).
  \item \textbf{Execution and measurement}: $\mathrm{Run}(c;\theta)$ executes a code artifact $c$ under evaluator configuration $\theta$ and returns a measurement record.
  \item \textbf{Selection rule}: either (i) a scalar utility mapping $U$ over measurement records, or (ii) a preference relation $\succ$ over measurement records. This rule may be implemented by automated metrics, test suites, rule-based comparators, an LLM-based judge, or a human-in-the-loop policy.
  \item \textbf{Stochastic aggregation}: $\mathrm{Agg}^R_K$ and $\mathrm{Agg}^J_K$ aggregate $K$ stochastic rollouts into an aggregated record and an aggregated utility estimate, respectively.
  \item \textbf{Stopping}: $\mathrm{Stop}(\beta, H)$ optionally terminates the loop based on budget progress and experiment history.
\end{itemize}

A \textbf{code artifact} $c \in \mathcal{C}$ denotes an executable repository state together with sufficient entrypoint/configuration to run under the evaluator. A single evaluator execution returns a measurement record
\[
R(c) = (\text{status},\; M(c),\; F(c),\; A(c)),
\]
where \texttt{status} indicates success or error, $M(c)$ denotes quantitative measurements (a scalar or vector of metrics), $F(c)$ denotes qualitative feedback (free-form text or structured diagnostics, including judge rationales), and $A(c)$ denotes auxiliary artifacts such as logs, traces, or produced files.

\subsection{Objective, feasibility, and stochastic evaluation}

KAPSO improves artifacts according to the evaluator's selection rule. In the common case, the evaluator provides a scalar utility mapping $U(R(c)) \in \mathbb{R}$. When execution is stochastic, the measurement record $R(c)$ is a random variable induced by evaluator randomness, and the conceptual optimization target is the \textbf{expected utility}
\[
J^\star(c) := \mathbb{E}\!\left[U(R(c))\right].
\]
This definition avoids ambiguity between $U(\mathbb{E}[R])$ and $\mathbb{E}[U(R)]$ when $U$ is nonlinear or when $R$ contains more than a single scalar.

KAPSO does not observe $J^\star(c)$ directly. Instead, the evaluator runs $K$ rollouts, where $K$ is part of evaluator configuration $\theta$. Let $R^{(k)}(c)$ be the record returned by rollout $k$. The evaluator defines an aggregation operator that produces an aggregated record
\[
\widehat{R}_{K}(c) := \mathrm{Agg}^R_K\!\left(R^{(1)}(c),\dots,R^{(K)}(c)\right),
\]
and an aggregation operator that produces an aggregated utility estimate
\[
\widehat{J}_{K}(c) := \mathrm{Agg}^J_K\!\left(U(R^{(1)}(c)),\dots,U(R^{(K)}(c))\right).
\]
In many instantiations, $\mathrm{Agg}^J_K$ is the sample mean and $\widehat{J}_{K}(c)$ is a Monte Carlo estimator of $J^\star(c)$.

The \texttt{status} field defines feasibility under the evaluator contract. Errors correspond to infeasible artifacts. The evaluator selection rule handles infeasibility by ensuring feasible artifacts are ranked above infeasible artifacts, for example by assigning a sentinel utility to errors in $U$ (such as $-\infty$ in a maximize setting) or by defining $\succ$ to always prefer feasible records over error records.

When an evaluator does not naturally provide a scalar utility, it instead provides a preference relation $\succ$ over (aggregated) measurement records, potentially using multi-metric and qualitative feedback. In that case KAPSO selects artifacts by comparing $\widehat{R}_K(\cdot)$ under $\succ$.

\subsection{Experiments, provenance, and history}

KAPSO maintains an explicit history of executed experiments. Each experiment corresponds to an isolated execution in an experiment branch and records the motivating specification, the produced artifact, and the measured outcomes. Let $b$ denote a branch identifier. We define the corresponding artifact as the repository state checked out from that branch:
\[
c := \mathrm{RepoState}(b).
\]

After iteration $i$, the experiment history is
\[
H_i = \{e_1,\dots,e_{i-1}\},
\]
with
\[
e_j = (b_j,\; u_j,\; \beta_j,\; c_j,\; K_j,\; \widehat{R}_{K_j}(c_j),\; \widehat{J}_{K_j}(c_j)),
\quad \text{where } c_j := \mathrm{RepoState}(b_j).
\]
Here $u_j$ is the solution specification used to generate edits for branch $b_j$, $\beta_j$ is the budget progress at which the experiment was executed, and $K_j$ is the rollout count used by the evaluator. When selection is defined purely by a preference relation $\succ$, $\widehat{J}$ can be omitted and KAPSO selects using $\widehat{R}$ directly.

\subsection{Knowledge grounding via seed repositories and retrieval}
\label{sec:formalization:seed_retrieval}

KAPSO grounds optimization in two complementary assets: (i) a \emph{seed repository} that provides an initial runnable codebase when available, and (ii) a typed knowledge graph that provides reusable principles, implementations, heuristics, and environment constraints as auxiliary context.

\paragraph{Repository corpus and seeding.}
Let $\mathcal{R}$ denote a corpus of candidate repositories (ingested via \texttt{learn} and indexed for retrieval). A repository $r \in \mathcal{R}$ includes its content plus metadata (e.g., tags, language, dependencies, and embedding-based representations). KAPSO uses a repository retrieval function
\[
\mathrm{RepoRetrieve}(g) \rightarrow (\hat{r}, \rho),
\]
which returns a candidate seed repository $\hat{r}$ and a confidence score $\rho \in [0,1]$. If $\rho$ exceeds a threshold $\tau$, KAPSO initializes the run from $\hat{r}$; otherwise it initializes from an empty (or template) scaffold:
\[
c_{\mathrm{init}} :=
\begin{cases}
\mathrm{RepoState}(\hat{r}) & \text{if } \rho \ge \tau,\\
\mathrm{Scaffold}(g) & \text{otherwise.}
\end{cases}
\]
Operationally, $c_{\mathrm{init}}$ becomes the parent state for the first experiment branch.

\paragraph{Typed knowledge graph and retrieval.}
Let $\mathcal{K}=(\mathcal{V},\mathcal{E}_K)$ be a typed knowledge graph where each node $v \in \mathcal{V}$ corresponds to a typed wiki page and has an ID, a title, a type in $\{\text{Principle},\text{Implementation},\text{Environment},\text{Heuristic}\}$, content text (and optionally code snippets), and typed edges in $\mathcal{E}_K$ (for example \texttt{IMPLEMENTED\_BY}, \texttt{USES\_HEURISTIC}, \texttt{REQUIRES\_ENV}, and cross-reference edges).

Given the goal $g$, an optional seed repository $\hat{r}$ (possibly $\emptyset$), and an optional failure signal $s$ from the latest experiment (e.g., error trace, contract violation, or qualitative evaluator feedback), KAPSO retrieves auxiliary knowledge via:
\[
\mathrm{RetrieveKnowledge}(g, \hat{r}, s) \rightarrow K.
\]

\paragraph{Base knowledge retrieval (KR).}
The base retrieval objective selects and returns a bounded set of relevant pages (principles, implementations, heuristics, and environments), optionally conditioned on the seed repository:
\[
K_{\mathrm{base}}(g,\hat{r}) := \mathrm{KR}(g,\hat{r}).
\]
Conditioning on $\hat{r}$ allows KAPSO to surface repo-specific constraints (dependencies, entrypoints, idioms) and to prioritize knowledge that is compatible with the starting codebase.

\paragraph{Error-recovery augmentation (ERA).}
After a failed experiment or repeated contract violations, KAPSO augments the retrieved packet with failure-conditioned heuristics, diagnostics, and alternative implementations:
\[
K :=
\begin{cases}
\mathrm{ERA}(g, s, \hat{r}, K_{\mathrm{base}}(g,\hat{r})) & \text{if } s \neq \emptyset,\\
K_{\mathrm{base}}(g,\hat{r}) & \text{otherwise.}
\end{cases}
\]

\paragraph{Knowledge packet.}
The output $K$ is a \textbf{knowledge packet} (implemented as \texttt{KGKnowledge}). It includes: (i) an optional seed repository reference $\hat{r}$ with confidence $\rho$, (ii) a set of retrieved pages grouped by type, (iii) confidence scores and provenance metadata (including \texttt{query\_used} and \texttt{source\_pages}), and (iv) optional recovery attachments when \textsc{ERA} is applied.

\subsection{Core solve loop (Orchestrator)}

At the top level, KAPSO repeatedly builds context and executes experiments until a stop condition fires. In the codebase this corresponds to \texttt{OrchestratorAgent.solve()} as invoked by \texttt{Kapso.evolve} plus a chosen \texttt{SearchStrategy}.

\begin{quote}
\textbf{Algorithm 1: KAPSO solve loop (orchestrator level)}
\begin{KapsoAlgo}
Inputs:
  - evaluator contract E = (P, Run, Stop, Select, Agg)
  - search strategy S (linear or tree)
  - context manager M
  - budgets B

Initialize i = 0
Initialize history H_0 = empty
while i < N:
  beta_i = budget_progress(B, i)
  if Stop(beta_i, H_i) or beta_i >= 1:
    break

  x_i = M.get_context(beta_i)
  # x_i includes P(beta_i), knowledge, episodic memory, and H_i

  new_experiments = S.run(x_i, beta_i)
  H_{i+1} = H_i union new_experiments
  i = i + 1

return best artifact in H_i under evaluator selection rule
\end{KapsoAlgo}
\end{quote}

This formalization makes explicit that KAPSO is a search over executable artifacts driven by repeated experiments and evaluator-defined assessment.

\subsection{Implement-and-debug loop (SearchStrategy)}

Both linear and tree search ultimately execute the same inner loop: create an isolated branch, implement a solution, run it, and optionally debug it for a bounded number of tries. The debug loop is intended to repair execution failures and evaluator-detected contract violations (for example, incorrect output format or missing required files), not to perform objective optimization.

\begin{quote}
\textbf{Algorithm 2: Implement-and-debug loop (branch level)}
\begin{KapsoAlgo}
Inputs:
  - solution spec u
  - context x (problem + knowledge + episodic memory + history)
  - debug budget D
  - branch name b and parent branch p

session = create_experiment_session(branch=b, parent=p)
r = implement_solution(u, x, session)   # codegen + Run

for k in 1..D:
  if r.has_error_or_contract_violation:
    r = debug_solution(u, x, r, session)
  else:
    break

finalize_session(session)               # commits + push + cleanup
return r
\end{KapsoAlgo}
\end{quote}

\subsection{LLM-steered tree search (one concrete instantiation)}

For completeness, we formalize an LLM-steered tree search instantiation used in some KAPSO configurations. Let $T=(\mathcal{N},\mathcal{A})$ be a tree of nodes $\mathcal{N}$, where each node stores a solution specification $u(n)$ and an optional experiment outcome derived from executing the corresponding artifact.

At each outer iteration, the strategy:
\begin{itemize}[leftmargin=1.2em]
  \item \textbf{Prune}: optionally terminates some leaf nodes using an LLM conditioned on $(P(\beta_i), K_i, H_i)$.
  \item \textbf{Expand}: chooses nodes to expand (exploration versus exploitation) and generates new child solution specifications via an LLM ensemble.
  \item \textbf{Select}: selects top-$k$ leaf nodes to execute as experiments, conditioned on the rendered context.
\end{itemize}

This can be viewed as a learned proposal distribution over solution specifications $u$ combined with black-box evaluation through $\mathrm{Run}(\cdot)$ and selection under $U$ or $\succ$.

\subsection{Cognitive memory (cascaded retrieval, episodic learning, decisions)}

KAPSO maintains an episodic memory store $E$ of reusable lessons extracted from experiment traces (run logs, diffs, and evaluator feedback). After each experiment, the system updates episodic memory, retrieves relevant prior lessons, and decides whether to continue iterating on the current workflow or pivot to alternative knowledge.

Let the controller state be $C_i$ containing the goal $g$, the current knowledge packet $K_i$, the latest experiment record $e_{i-1}$, retrieved episodic insights $Z_i$, and meta statistics (for example, consecutive failures).

The controller defines:
\begin{itemize}[leftmargin=1.2em]
  \item $\mathrm{RetrieveCascade}(g, s)\rightarrow K_i$: cascaded retrieval using \textsc{WSR} with \textsc{PFR} fallback, plus \textsc{ERA} augmentation when $s$ indicates failure or contract violation,
  \item $\mathrm{UpdateEpisodic}(e_{i-1})\rightarrow E$: store generalized lessons from errors and qualitative feedback,
  \item $\pi(C_i)\in\{\textsc{Retry},\textsc{Pivot},\textsc{Complete}\}$: a policy over iteration-level actions.
\end{itemize}

\begin{quote}
\textbf{Algorithm 3: Cognitive controller step (per experiment)}
\begin{KapsoAlgo}
Inputs: goal g, current knowledge K, episodic memory E, last experiment e

if e.has_error_or_contract_violation:
  E.add(ExtractIssue(g, e))
  K = ERA(g, e, K)   # recovery heuristics + alternatives with provenance
else if e.feedback is non-empty:
  E.add(ExtractInsight(g, e.feedback))

Z = RetrieveEpisodic(E, g, e)
a = DecideAction(g, K, e, Z)   # RETRY / PIVOT / COMPLETE

if a == PIVOT:
  K = RetrieveCascade(g, s=None)   # exclude current workflow in implementation

return (a, K, Z)
\end{KapsoAlgo}
\end{quote}

The key property is that both the coding agent and the decision maker are grounded in the same rendered context, and ERA explicitly records provenance via \texttt{query\_used} and \texttt{source\_pages}.

\section{System Implementation}
\label{sec:implementation}

This section describes the implementation of the mechanisms defined in Section~\ref{sec:formalization}. The key design principle is that framework semantics (artifact, evaluator contract, experiment history, seed repository selection, knowledge retrieval, and controller loops) are fixed, while concrete implementations of execution, storage, indexing, and adapters remain modular.

\subsection{Experimentation Engine}
\label{sec:implementation:experiments}

The experimentation engine is the mechanism that realizes KAPSO's experiment history and provenance model from Section~\ref{sec:formalization}. Its primary role is to isolate attempts, make outcomes reproducible, and enable reuse of successful attempts as parents for subsequent experiments.

\subsubsection{Experiment sessions as git branches}

KAPSO represents each experiment as a git branch inside an \texttt{ExperimentWorkspace}. \\An \texttt{ExperimentSession} starts from a parent branch, creates a new branch identifier $b$, applies edits using the configured \texttt{CodingAgent}, and executes the evaluator through the active \texttt{ProblemHandler}. The session then commits the resulting artifact state and run outputs to the branch. This makes each experiment concrete and inspectable: a branch can be checked out and re-executed to reproduce the same evaluator interaction and artifacts.

To support downstream reuse, \texttt{ExperimentSession.close\_session()} always attempts \texttt{git push origin <branch>}. In typical deployments, \texttt{origin} is not a network remote. Sessions are cloned from \texttt{file://<base\_repo.working\_dir>}, where \texttt{base\_repo} is the initialized parent artifact for the run (either a retrieved seed repository or a scaffold). Pushing primarily publishes the branch back into the local \texttt{ExperimentWorkspace} repository. This ensures that newly created branches are immediately available as parents for child experiments in tree-based exploration. It is not guaranteed to publish to GitHub or any external remote.

Each experiment persists a lightweight, reproducible bundle sufficient for debugging and audit. Concretely, the committed artifacts include:
\begin{itemize}[leftmargin=1.2em]
  \item the code changes relative to the parent branch (diffs and commit identifiers),
  \item evaluator configuration used for the run (including rollout count $K$ when stochastic aggregation is enabled),
  \item run logs and structured diagnostics, and
  \item evaluator-produced artifacts (for example output files and traces).
\end{itemize}

\subsubsection{Execution environment and scaling}

Experiments execute inside the active \texttt{ProblemHandler}. In the default path, \texttt{GenericProblemHandler} runs the artifact via local subprocess, and no containerization is introduced by the experimentation engine itself. Containerized execution is used only when the evaluator requires it. For example, ALE evaluation is performed by an external evaluation harness that runs solutions in Docker, while MLE runs the Python entrypoint locally inside the benchmark environment. This division keeps the experimentation engine evaluator-agnostic while still supporting evaluators with strict runtime requirements.

The engine supports executing multiple \texttt{ExperimentSession}s concurrently, which is used by search strategies that evaluate multiple candidates in parallel. While the current implementation primarily runs locally (or via evaluator-required containers), the design admits a pluggable execution backend. In particular, the same session and artifact model can be paired with a remote executor to run resource-intensive workloads (for example GPU-heavy training) on appropriate machines, while preserving the branch-based provenance that downstream search and reuse depend on.

\subsection{Knowledge System}
\label{sec:implementation:knowledge}

The knowledge system converts heterogeneous sources into (i) a repository corpus used for initialization and reuse, and (ii) a typed knowledge base indexed for retrieval. MediaWiki provides a familiar interface for human review, curation, and editing, while the same content is served to agents through retrieval APIs. The retrieval semantics---seed repository selection plus typed knowledge retrieval with failure-conditioned augmentation (ERA)---are defined in Section~\ref{sec:formalization:seed_retrieval}. This subsection describes how those semantics are implemented and how repositories and knowledge are acquired, represented, and indexed.

Knowledge acquisition is orchestrated by a \texttt{KnowledgePipeline} that produces typed wiki pages (\texttt{WikiPage}) and maintains a \texttt{RepoStore} containing versioned references to ingested repositories. For repositories, \texttt{RepoIngestor} follows a repository-centric pipeline designed for two outputs: (i) a runnable seed codebase (the repository snapshot itself, referenced by URL and commit metadata), and (ii) structured knowledge extracted from that repository for reuse across tasks. Concretely, the ingestor identifies entrypoints, dependency and environment requirements, and high-signal implementation patterns; extracts reusable principles, implementations, heuristics, and environment constraints; and attaches provenance linking each extracted item back to repository paths and commits. In addition, the pipeline performs deterministic file mining to surface useful files that are not emphasized in READMEs (for example scripts, configs, evaluation utilities, deployment manifests), triages them, and emits corresponding typed pages when they contain reusable guidance.

Knowledge is represented as typed pages with explicit, typed links. KAPSO uses four primary page types: \texttt{Principle}, \texttt{Implementation}, \texttt{Environment}, and \texttt{Heuristic}. Links between pages become typed edges (for example \texttt{IMPLEMENTED\_BY}, \texttt{USES\_HEURISTIC}, and \texttt{REQUIRES\_ENV}). Extracted pages carry repository provenance and optional repository associations (for example \texttt{SOURCE\_REPO} metadata and \texttt{RELATED\_REPO} links) to enable repo-conditioned retrieval. This structure is designed to return bounded, typed knowledge packets that can be consumed by the evolve loop without requiring workflow synthesis.

For machine retrieval, the wiki and repository corpus are indexed into pluggable backends. The reference implementation ships with a typed graph index (default Neo4j) and a vector index (default Weaviate), while allowing alternative graph stores and vector databases behind the same interfaces. Retrieval is implemented in two stages. First, a \texttt{RepoRetrieve} service selects an optional seed repository by searching over the repository corpus using hybrid signals (for example embeddings over READMEs and file trees, plus metadata filters) and returns a candidate seed with a confidence score. Second, a \texttt{KnowledgeRetrieve} service retrieves relevant principles, implementations, heuristics, and environment constraints from the typed indices, optionally conditioned on the selected seed repository. After failures or repeated contract violations, the retrieval service applies error-recovery augmentation (ERA) to attach failure-conditioned heuristics and alternative implementations. Across all modes, the system records provenance into the returned knowledge packet (including \texttt{query\_used}, \texttt{source\_pages}, and optional seed repository identifiers) to support auditability and reproducibility.

We release a complete knowledge package consisting of a MediaWiki dump, Neo4j and Weaviate snapshots, and Docker-based deployment scripts that bring up the MediaWiki instance and all indices in a reproducible configuration. The released package also includes a manifest of the repository corpus used for seeding and extraction (for example URLs, commit identifiers, and selection criteria), enabling point-in-time reconstruction of the repository set.

\subsection{Cognitive Memory System}
\label{sec:implementation:memory}

The cognitive memory system implements episodic memory and controller decisions as formalized in Section~\ref{sec:formalization}. Its goal is to reduce repeated error modes, preserve high-value lessons from prior attempts, and make long-horizon iteration more stable by conditioning future proposals and debugging on prior experience.

Episodic memory entries are derived from experiment traces and are designed to be reusable across tasks. Each entry includes a compact trigger description, a generalized lesson, recommended actions, and provenance linking back to the originating experiment branch and its artifacts (for example logs, diffs, and evaluator feedback). This provenance makes the memory auditable and supports inspection when a retrieved lesson influences a new change.

After each experiment, KAPSO performs lesson extraction. When the experiment indicates an execution error or contract violation, the system generalizes the failure into a reusable fix pattern grounded in the observed trace and validator feedback. When the experiment succeeds or improves quality, the system extracts best-practice insights from measured outcomes and qualitative feedback, including judge rationales when an LLM-based evaluator is used. Extracted lessons are stored in a vector database (default Weaviate) with a JSON fallback, enabling semantic retrieval by goal and by failure signal.

On subsequent iterations, KAPSO retrieves relevant episodic memories conditioned on the current goal and the latest experiment signal. Retrieved lessons are rendered into the unified context passed to the search strategy and coding agent, alongside the current knowledge packet produced by repository selection and typed knowledge retrieval. This ensures that both proposal and debugging steps are grounded not only in domain knowledge (the wiki and indices) but also in the system's own prior experience on similar issues.

Finally, the controller implements an iteration-level decision policy as in Algorithm 3 in Section~\ref{sec:formalization}. It uses the goal, the current knowledge packet, the latest experiment record, and retrieved episodic insights to decide whether to retry, pivot, or complete. On pivot, the implementation re-runs repository retrieval and knowledge retrieval while excluding the currently selected seed repository (when one was used), encouraging exploration of alternative starting codebases or a fallback to an unseeded scaffold.

\subsection{Deployment Interface}
\label{sec:implementation:deploy}

KAPSO provides a unified deployment interface that packages a selected solution artifact into a runnable form while preserving a stable invocation contract. The user-facing contract is described in Section~\ref{sec:framework:deploy}; this subsection describes the implementation mechanics that realize that contract across deployment strategies.

\texttt{Kapso.deploy(...)} returns a Python object implementing the \texttt{Software} interface. The returned handle exposes a stable \texttt{run(inputs)} method and lifecycle methods (for example \texttt{start}, \texttt{stop}, \texttt{logs}, and \texttt{is\_healthy}). The key implementation goal is to keep this contract invariant while allowing strategy-specific execution details to vary.

Deployment is done via repository adaptation. Given a selected solution repository, it creates an adapted copy at a path of the form \texttt{<solution.code\_path>\_adapted\_<strategy>}. The adapter injects strategy-specific runtime wrappers (for example, container or service scaffolding or platform configuration) and emits a run interface descriptor that specifies how to invoke the artifact (for example an endpoint URL, a module path, or a callable reference). The \texttt{Software} handle uses this descriptor to route \texttt{run()} to the appropriate local or remote mechanism.

The current implementation supports multiple strategies under the same \texttt{Software} interface. In \textbf{LOCAL} mode, the runner imports and calls a function inside the adapted repository (default \texttt{main.predict}). In \textbf{DOCKER} mode, the system builds or reuses a Docker image and runs a local container exposing an HTTP endpoint (default \texttt{http://localhost:8000/predict}). In \textbf{MODAL} mode, it generates a \texttt{modal\_app.py} and invokes a remote Modal function. In \textbf{BENTOML} mode, it generates BentoML service files and can optionally deploy to BentoCloud, returning an HTTP endpoint. In \textbf{LANGGRAPH} mode, it generates LangGraph deployment files and the runner connects to a LangGraph Platform URL to invoke the deployed agent. Across all strategies, the caller interacts only with \texttt{Software.run()}, while lifecycle methods expose health and logs in a strategy-appropriate way.

The deployment layer is extensible. New strategies can be added by implementing an adapter that (i) produces the required runtime wrapper files, (ii) emits the run interface descriptor, and (iii) registers how the \texttt{Software} handle should execute \texttt{run()} and lifecycle methods for that strategy.

\section{Evaluation}

We evaluate KAPSO on two benchmarks that capture distinct real-world software-building regimes.

\subsection{MLE-Bench (Kaggle-style ML competitions)}
\textbf{Task}: produce a Python solution that trains on provided competition data and writes a \path{final_submission.csv} in the expected format.
\\
\\
\textbf{Execution protocol} (as implemented in the benchmark handler):

\begin{itemize}[leftmargin=1.2em]
  \item run \textbf{debug mode} first (\texttt{\detokenize{python main.py --debug}}) with a strict runtime cap,
  \item validate the submission file format,
  \item run \textbf{full mode} (\texttt{\detokenize{python main.py}}) with a longer runtime budget,
  \item grade the output submission on the competition's train/test split of public train dataset.
\item \textbf{Stop condition}: Run for 24 hours or until a maximum budget of \$200 is reached. Also, stop early if the run achieves \textbf{any medal} according to the MLE-Bench grading library.

\end{itemize}

\noindent
\textbf{Reported metrics}:

\begin{itemize}[leftmargin=1.2em]
  \item \textbf{Private score}: returned by the benchmark grader.
  \item \textbf{Medal rate}: fraction of competitions where any medal is achieved in 4 categories: low, medium, hard, and all.
\end{itemize}

\noindent
\textbf{Results}:

Table \ref{tab:mlebench} indicates that Leeroo consistently outperforms leading open-source agent frameworks, and its advantage becomes more evident as the difficulty of tasks increases. Although performance in Low-difficulty tasks is identical between Leeroo and the top-tier open-source baseline, R\&D-Agent (both achieving 68.18\%), Leeroo achieves substantially higher accuracy on Medium and Hard problems, reaching 44.74\% and 40.00\%, respectively. In comparison, the highest-performing open-source agent, R\&D-Agent, attains only 21.05\% on Medium and 22.22\% on Hard tasks. These results suggest that while open-source scaffolding is competitive on standard problems, Leeroo’s capabilities transfer more effectively to complex machine learning challenges involving high levels of specialization and long-horizon engineering.

\begin{table}[ht]
\centering
\begin{tabular}{l c c c c}
\hline
Agent & Low (\%) & Medium (\%) & Hard (\%) & All (\%) \\
\hline
Leeroo & \textbf{68.18 $\pm$ 2.62} & \textbf{44.74 $\pm$ 1.52} & \textbf{40.00 $\pm$ 0.00} & \textbf{50.67 $\pm$ 1.33} \\
R\&D-Agent\cite{rd-agent} & 68.18 $\pm$ 2.62 & 21.05 $\pm$ 1.52 & 22.22 $\pm$ 2.22 & 35.11 $\pm$ 0.44 \\
AIRA-dojo\cite{aira} & 55.00 $\pm$ 1.47 & 21.97 $\pm$ 1.17 & 21.67 $\pm$ 1.07 & 31.60 $\pm$ 0.82 \\
ML-Master & 48.48 $\pm$ 1.52 & 20.18 $\pm$ 2.32 & 24.44 $\pm$ 2.22 & 29.33 $\pm$ 0.77 \\
AIDE & 35.91 $\pm$ 1.86 & 8.45 $\pm$ 0.43 & 11.67 $\pm$ 1.27 & 17.12 $\pm$ 0.61 \\
\hline
\end{tabular}
\caption{Performance comparison of Leeroo against the top four open-source agent frameworks on MLE-bench.}
\label{tab:mlebench}
\end{table}

See the official \href{https://github.com/openai/mle-bench}{MLE-Bench repo} for the most up-to-date results.

\subsection{ALE-Bench (AtCoder heuristic contests)}

\textbf{Task}: produce a C++ (\texttt{cpp23}) solution in \texttt{\detokenize{main.cpp}} to maximize/minimize a contest-defined score under a strict runtime limit of each competition.

\vspace{0.75em}

\noindent
\textbf{Execution protocol}:
\begin{itemize}[leftmargin=1.2em]
  \item compile and run in the benchmark's Docker environment,
  \item if the solution is accepted, run it multiple times and average the public evaluation score to reduce the effect of randomness.
  \item Run the private evaluation for the experiment with the highest public score and report final performance and rank percentile.
\end{itemize}

\noindent
\textbf{Reported metrics}:

\begin{itemize}[leftmargin=1.2em]
  \item \textbf{Final performance}: Final ELO rating of the submission for the competitions.
  \item \textbf{Rank percentile (RP)}: \texttt{final\_rank} / \texttt{number\_of\_contestants}, Compares the agent’s performance with the human-level performance.
  \item \textbf{Private absolute score} (contest objective value).
  \item \textbf{Cost}: cumulative LLM usage cost.
\end{itemize}

\noindent
\textbf{Results}:

Table \ref{tab:alebench} shows that Leeroo achieves the highest final performance on the ALE Bench, scoring 1909.4 with a rank percentile of 6.1\%, outperforming the original ALE-Agent (1879.3, 6.8\%). Notably, Leeroo does this while keeping total cost relatively low at \$914.8, compared to ALE-Agent’s \$1003.3, indicating a more cost-efficient approach. The competition-level results in Table \ref{tab:ahc_grouped} further highlight Leeroo’s advantage: it surpasses ALE in most AHC competitions, sometimes by large margins—e.g., ahc016 (2022 vs. 1457) and ahc026 (2040 vs. 1965)—while maintaining comparable rank percentiles. These outcomes demonstrate that Leeroo not only achieves higher absolute performance but also generalizes more reliably across diverse competitions, combining effectiveness with efficiency, which makes it a more impactful agent for real-world ALE tasks.

One limitation of the current results is that the relatively low number of competitions in ALE-Bench may introduce noise, as the performance of LLMs and agents can vary across tasks. For example, in competition ahc039, ALE-Agent achieves a notably high score, but this performance is not consistently reflected in other similar short competitions. Future studies including a larger set of competitions, running with multiple seeds, could provide a more robust and reliable comparison between agents.

\begin{table}[h]
\centering
\begin{tabular}{l c c c c c}
\hline
Agent  & Final Performance & Rank Percentile (\%) & Total Cost (\$) & Model \\
\hline
Leeroo & \textbf{1909.4} & \textbf{6.1}& 914.8 & Gemini-2.5-pro \\
ALE-Agent \cite{ale-agent} & 1879.3 & 6.8 & 1003.3 & Gemini-2.5-pro \\
ALE Sequential \cite{ale-agent} & 1198 & 54.1 & 111.0 & Gemini-2.5-pro \\
ALE one-shot \cite{ale-agent} & 832 & 88.4 & 4.7 & Gemini-2.5-pro \\

\hline
\end{tabular}
\caption{Aggregated results on ALE Bench.}
\label{tab:alebench}
\end{table}

\begin{table}[h]
\centering
\begin{tabular}{l cc cc}
\hline
Competition & \multicolumn{2}{c}{ALE} & \multicolumn{2}{c}{Leeroo} \\
\cline{2-5}
 & Final Performance & RP (\%) & Final Performance & RP (\%) \\
\hline
ahc008 & 1189 & 52.06 & 1221 & 49.03 \\
ahc011 & 1652 & 20.30 & 1607 & 22.35 \\
ahc015 & 2446 & 3.85  & 2528 & 2.82  \\
ahc016 & 1457 & 33.05 & 2022 & 9.17  \\
ahc024 & 1980 & 13.10 & 1980 & 13.10 \\
ahc025 & 1331 & 47.00 & 1353 & 46.30 \\
ahc026 & 1965 & 16.00 & 2040 & 12.43 \\
ahc027 & 1740 & 18.12 & 1839 & 14.91 \\
ahc046 & 2153 & 8.95  & 2194 & 8.09  \\
ahc039 & 2880 & 0.73  & 2310 & 5.27  \\
\hline
\end{tabular}
\caption{Comparison of ALE and Leeroo performance across AHC competitions.}
\label{tab:ahc_grouped}
\end{table}

\newpage

\section{Conclusion}
KAPSO is a framework for building software by running evaluator-grounded experiments, using structured knowledge, and learning from episodic experience. Its core contributions are a git-native experimentation engine, a scalable knowledge acquisition and retrieval system with cascaded retrieval (WSR with PFR fallback, plus ERA augmentation), and a workflow-aware cognitive memory layer that stores and reuses lessons from experiment traces. The framework is designed to be modular and auditable, with a clear evaluator contract and reproducible artifacts. We provide a clear execution protocol for MLE-Bench and ALE-Bench to enable reproducible evaluation.

\appendix

\end{document}